# TOC

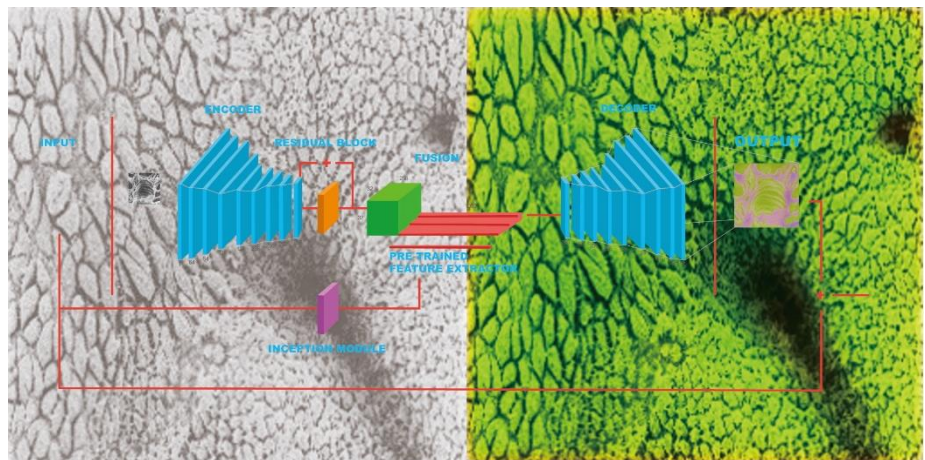

Giving grayscale microscopy images a colorful world by using artificial intelligence technology.

# Nanoscale Microscopy Images Colorization Using Neural Networks

Israel Goytom[1,4] *, Qin Wang[2] *, Tianxiang Yu[1], Kunjie Dai[1], Kris Sankaran[4], Xinfei Zhou[1],

Dongdong Lin[1,3]

1 Department of Microelectronics Science and Engineering, Ningbo University, Ningbo,

China 315211

2 Shenzhen Research Institute of Big Data, The Chinese University of Hong

Kong(Shenzhen), Shenzhen, China 510275

3 Department of physics, Fudan University, Shanghai, China 200433

4 Montréal Institute for Learning Algorithms (MILA), Montréal Canada, H2S 3H1

* Israel Goytom and Qin Wang contributed equally

Corresponding author: lindongdong@nbu.edu.cn

# ABSTRACT

Microscopy images are powerful tools and widely used in the majority of research areas, such as biology, chemistry, physics and materials fields by various microscopies (scanning electron microscope (SEM), atomic force microscope (AFM) and the optical microscope, et al.). However, most of the microscopy images are colorless due to the unique imaging mechanism. Though investigating on some popular solutions proposed recently about colorizing images, we notice the process of those methods are usually tedious, complicated, and time-consuming. In this paper, inspired by the achievement of machine learning algorithms on different science fields, we introduce two artificial neural networks for gray microscopy image colorization: An end-to-end convolutional neural network (CNN) with a pre-trained model for feature extraction and a pixel-to-pixel neural style transfer convolutional neural network (NST-CNN), which can colorize gray microscopy images with semantic information learned from a user-provided colorful image at inference time. The results demonstrate that our algorithm not only can colorize the microscopy images under complex circumstances precisely but also make the color naturally according to the training of a massive number of nature images with proper hue and saturation.

Keywords: Colorization, Deep learning, Microscopy images, Neural networks

# 1 Introduction

Nanotechnology is the art and science of manipulating matter at nanoscale to create new and unique materials and products [1-4]. Electron microscopes like the scanning electron microscope (SEM) [5, 6] and transmission electron microscope (TEM) [7], near-field microscopes like atomic force microscope (AFM) [8, 9] and scanning tunnelling microscope (STM) [10, 11] are widely used in various scientific fields. As they are versatile imaging and measurement sources, they help researchers and scientists visualize structures from nanoscale or microscale level, providing observations required for nanoscience and nanotechnology experiments [12]. These imaging techniques are strongly essential to various types of researchers.

Since some of these microscope devices, form images with physical parameters instead of photons of visible light. For example, SEM produces images by scanning the surface of the sample with a focused beam of electrons, while AFM measures the forces between the probe and the sample as a function of their mutual separation. Hence, the resulting images are grayscale, and these single-channel images contain less information than the colorized version of the same image. Despite the many advantages of imaging results by these devices, we do not see the full micro-world. For example, most of the micro images contain multi-components structures, however, there are only gray image obtained by SEM /TEM/AFM, and so on. As a result, using the grayscale images out-of-the-box may result in limited observation between the experimenter and their audience. To prevent these limitations and to help bridge the gap, researchers have developed different tools, and authors spend money and time to redraw or to colorize their experimental images as polychrome, while faithfully representing the precise 3D appearance of the samples.

Colorization of grayscale SEM images is a common task among researchers working with microscopy. In most images, the signal intensity at each pixel corresponds to a single number that represents the proportion number of electrons emitted, and these numbers are usually described as a grayscale image. Perhaps the most common approach is to use software like Mountains-Map, Adobe Photoshop, Fiji, Image J, or even by-hand. This is time-consuming and requires distinctive knowledge. Some methods come with software tools for automatic colorization [13], but they still require massive user inputs. This process requires extensive background researches and inputs, just to colorize one single image. The process might start from reading or researching about what is in the image and how to represent the color, ending with restoring the original content and colorizing the arrangement. In some scenarios, a picture can take up to several days to colorize, and it requires extensive research if the structures in the image are complicated. Some objects might need multi-layers and different color shades to get it just right; it requires specialized knowledge of software or manual skills. Furthermore, gradient ramp and fusion between two colors are extensively common in micro-world images, and it is a huge obstacle for ordinary software and hand drawing.

To solve these problems, previously, abundant colorization methods were developed and worked well. Some of the methods give an input grayscale image and dene one or more color reference images (provided by a user or retrieved automatically) to be used as source data. Then, following the image analogies framework, color is transferred onto the input image from analogous regions of the reference image [14, 15]. Parametric methods, on the other hand, learn prediction functions from large datasets of color images at training time, posing the problem as either regression onto continuous color space [16, 17] or classification of quantized color values [18]. Researchers proposed a multi-modal scheme, where each pixel was given a probability for each possible color [19]. A recent method introduced a model which combines a deep convolutional neural network (D-CNN) trained from scratch with high-level features extracted from a pre-trained model [20]. There are also methods which leverage large-scale data and CNNs [21, 22]. In terms of loss function of the methods, classification loss was used with rebalanced rare classes [19], other works use an un-rebalanced classification loss [21] and a regression loss [22]. However, these methods are for general-purpose image colorization and are not explicitly designed for nano/micro-structural microscopy images. To our knowledge, most colorization works on microscopy images, and their image colorization used graphics software and human hand drawing, not deep learning methods.

In this paper, we proposed two approaches (end-to-end convolutional neural network (EE-CNN) and NST-convolutional neural networks models (NST-CNN)) to colorize the grayscale microscopy images by using artificial neural networks. These algorithms learn multiple levels of color representations corresponding to different levels of abstraction in microscopy images. To improve the relationship between the samples under the microscopes and their real-world representative, and to answer what will help our algorithm understand what is in the image and their relations to one another, a pre-trained Inception-ResNetV3 model is exploited as a feature extractor. This pre-trained model was trained on existed dataset ImageNet [23]. Microscopy images might not have one exact color or ground truth to compare with the output. Instead, colorizing these images must look realistic and represent their desired appearance. By our designed models, the results demonstrated a good colorization ability on most microscopy images with/without substrate.

## 2 Experimental and methods

Our models are based on CIELAB [24] color space which exploits "*LAB*" color-opponent space to represent color instead of "*RGB*" space. Dimension "*L*" indicates luminance of the image. Dimension "*A*" and "*B*" represent the red-green and blue-yellow color spectrum respectively. More Specifically, for each given image, it has a brightness channel "*L*" and color channel "*A*" and "*B*". The proposed methods use neural networks to predict "*A*" and "*B*" color and contrast values from a brightness input "L", since grayscale images can be considered pixel values with the only "*L*" in the "*LAB*" color space. By using "*LAB*" color space, our model will only learn how to colorize the images without modifying light intensities. Predicting the "*AB*" in "*LAB*" color space can be either trained using colorful microscopy images or informed by a colorful reference photo, which corresponds with two proposed methods respectively. Finally, the output images are converted into red, green and blue in "*RGB*" color spaces to use them in general application. The full systematic architecture has pre-process and post-process steps to handle all the input and output images before and after colorization. The pre-processing mainly takes the resizing and reshaping tasks, while the post-processing improves the performance by a "same gray same color" constraint.

## 2.1 Dataset

Dataset is collected in this paper, to the best of our knowledge, this is the first colorful SEM images dataset. The dataset was a collection of ~ 1000 colorful SEM images (each image in this dataset is an "*RGB*" image), and it was divided as 90% for train set and 10% for the test. The images have different shapes and sizes due to varying resolution of the images; for simplicity, all images in the dataset were resized to the resolution of 300×300. During the training time, these images can be an input for the network to extra features to predict the color representation.

## 2.2 EE-CNN model

We proposed an end to end convolutional neural network (EE-CNN) to colorize the gray images. The network is able to yield "*AB*" color channel of an image from an input "*L*" channel value. EE-CNN was fully trained on "SEMCOLOURFUL" dataset which contains around 1000 colorful microscopy images. However, deep learning based methods usually require large number of samples to achieve a high performance. To address the lack of data issue, instead of training a feature extractor from scratch, a pre-trained classifier model Inception-ResNetV3 [26] was utilized to alignment the semantic information between microscopy samples and their real-world colorful representative, since the pre-trained model was trained on ImageNet with over 1.2 million images. More specifically, it can extract the semantic features from input image by getting the output of the final layer of the pre-trained model. The features can be used as an auxiliary input for latter encoder-decoder part to improve the performance and accelerate the convergence of the training process.

Our Proposed EE-CNN is shown in Figure 1(a). The model has four components:

- Encoder: Taking the input gray image and yields semantic features from the input.
- Inception module (Feature extractor): A pre-trained model to extract more semantic information to enhance the performance.
- Fusion layer: Merging features from Encoder and Inception module together.
- Decoder: Based on the merged features to reconstruct the content and select the appropriate color adaptively to generate the predict "*AB*" channels prediction.

**Encoder*:*** The encoder process is taking $H \times W$ images with "*L*" channel to produce outputs of $\frac{1}{8}(H \times W) \times 512$ features as show in encoder part (Figure 1(a)). It has 8 convolutional layers. The kernel size of each layer is 4 × 4. Stride of each odd numbered layers is halved, which is able to reduce its size and distill features.

**Inception module**: The extractor is a pre-trained model which is trained on a large benchmark dataset by conducting a classification task. For our task, due to the computational cost of training and lack of data. In this paper, instead of training feature extraction from scratch, we take merits from transfer learning.

**Fusion** : The fusion method states as follows, by taking a feature vectors extracted from encoder to repeat $\frac{H}{8} \times \frac{W}{8}$ times and then combined it with the features extracted from the inception module as the input of fusion layer. The fusion layer is a convolution layer with kernel size 4×4 [23] as show in the fusion part of Figure 1(a). By mirroring the feature vector and concatenating the vector several times it will help to ensure that the semantic information conveyed by the feature vector is uniformly distributed among all spatial regions of the image.

**Decoder**: The decoder takes the output from the fusion layer and apply convolutional and up-sampling operations to get the final image of size $H \times W \times 2$ which is the same size with the real values of "*AB*" color channels. The up-sampling part was performed using nearest neighbour approach.

The summarised flow of this method is by providing the luminance component ("*L*") of image. The model estimates the remaining "*AB*" color components from the input "*L*" and combines it with the predicted "*A*" and "*B*" to form an output. Finally, the output image is converted to "*RGB*" color space which is frequently used in computer system.

## 2.3 NST-CNN model

Inspired by a phenomenon that some microscopy images, like red blood cells, plants leave images have known colors and the shapes of many microscopy images are very similar to some natural images such as the shape of grapes and cells. we propose a method to predict the "*AB*" channels of a microscopy image from a colorful natural image. Concretely, we refer to recent works in neural style transfer (NST) [27] and propose a novel method called neural style transfer based on convolutional neural network (NST-CNN). The input grayscale microscopy image is treated as content image. A colorful natural image is utilized as reference image to provide color dye scheme for content image. Different from traditional neural style transfer, in our method, the microscopy image only receives color information as the style rather than shapes. Therefore, the model can transfer color precisely from reference image to content image without any change on content or shape.

The pipeline of NST-CNN is for pixel-to-pixel color transfer as shown in Figure 1 (b). The model takes "*L*" channel of reference image ($X^R$) as the input to yield the color channels ($X^{AB}$). Hence, the task is supervised similarly like EE-CNN method but only learning from single one reference image. Once the model is well optimized, we can do inference on the content image ($X^L$) with converged network weights which had been well updated in the training phrase. Therefore, the learned color will be transfer into the content image to colorize adoptively. In the post-process, the predicted "*AB*" channels results are optimized by following the same "*L*" and same "*AB*" constraint. Concretely, we take the value of "*L*" channel from the input image and find the pixels which have same "*L*" value will be assigned with same "*AB*" value. Hence, it can guarantee that the pixels with same "*L*" value will be assigned similar "*AB*" color to mitigate the miss-colorize issue.

To summarize the process, first, our NST-CNN model utilizes the natural reference image to fit the model to learn colorization information in a semantic way. Then the well optimized model can be adopted to colorize the "similar-content" gray microscopy images directly.

**2.4 Loss function**

Both EE-CNN and NST-CNN are optimized by mean squared error loss which is computed pixel by pixel between the estimated "*A*" and "*B*" channels and their corresponding ground truth which is their real value extracted from original image.

$$L(Y_{ab}, Y) = \frac{1}{h \times w} \sum_{\substack{0 \leq i < h \\ 0 \leq j < w}} \left(Y_{ab}^{i,j} - Y^{i,j}\right)^2$$

Where $Y_{ab}$ and $Y$ are the predicted and the real values of "*AB" channels* respectively, and $h$ and $w$ are the height and width of the images. The Adam optimizer [28] back-propagates the gradients computed from the loss function. To allow batch processing, the input image size is fixed during the training time.

**2.5 Training process**

The network trained and tested on MILA's cluster node, leveraging the NVIDIA CUDA 10 Toolkit, and the 3 NVIDIA Tesla K80 GPU for training. The full training took around 6.5 hours with the batch size of 16 to avoid overflowing of the GPU memory. Adam optimizer [28] is used during the training process; the initial learning rate is 0.0001. The loss function is mean squared error loss calculated in equation. The model was trained for 300 epochs and early stop to avoid the overfitting problem [29]. The whole program was written using the python programming language, and we used Tensorflow [30] and Keras [31] deep learning framework libraries.

## 3 Results and discussion

Different microscopy images scenarios were checked and then presented the approach on SEM micrograph of human cell, plant tissues, pollen grains, fab-nanostructure, insects, or shells and other commonly used specimen microscopy samples (the source links of grayscale images from internet are shown in supporting List 1). The results from the colorizing model vary in the different samples under microscope scenarios. In various situations, our algorithm performed well in colorization task.

### 3.1 Colorization by EE-CNN model

Our end-to-end convolutional neural network (EE-CNN) approach is designed to identify an input microscopy images and colorize them independently and correctly. Most microscopy images are different and might not have ground truth or one true color that we need to identify them accurately. To determine the objects and their relation with the real-world objects, The pre-trained model is applied to find the patterns and give them the color. Figure 2 illustrates the results from our EE-CNN model. Several grayscale microscopies images (bacillus, cilia, wood surface, nuclear membrane, stamens, leaf surface) are selected as the research samples (the web links of images are shown in Table S1). It is obviously that the colorization results display a beauty of multi-color with soft color gradient (Figure 2 ((h)-(m))). For example, the detailed components in Figure 2(c) could be easily distinguished in our colorized image (Figure 2(j)). In addition, the combination of luminosity and color in colorized images dramatically improve the quality of images (Figure 2(k), (m)). Significantly, the images of leaf surface can be recognized precisely as the pre-trained dataset contain huge number of nature images.

To check the colorized results of our EE-CNN model, the distribution of each color composition is analysed. "*HSV*" (hue, saturation, value) are alternative representations of the "*RGB*" color model [32]. The selected saturation value in Figure 2(n)) and Figure 2(o) are obtained from Figure 2(l) and Figure 2(h), respectively. The colorized results show a relatively uniform saturation intensity distribution in main objects (bacillus, stamens). Furthermore, another two type of images were colorized by our algorithm. Pollen granule image without substrate in Figure S2(a) and barium carbonate crystals grow on substrate in Figure S2(d). The EE-CNN model paints the green color to the pollen granule, and gold color to barium carbonate crystals. Form the separation of "*L*" and "*AB*" channels in output images, it clearly shows that the color matches to the pollen particles well (Figure S2(b)). In terms of barium carbonate crystals with substrate, the crystals could be colorized properly (Figure S2(e)). However, the substrate was also colorized by gold value, together with the influence of the shadows.

A pre-processing strategy is applied for the keeping of uniform color on one kind of object in an image to improve the quality of colorization. Holistically nested edge detection [33] is used to obtain the edge from input images Figure 3((b), (e)) before the colorization. Based on those boundary condition, the model colorizes the bounded area with a same color, this process is shown in Figure 3(c, f). However, this pre-processing strategy is successfully applied in the strong contrast images but failed (not well) in some soft edge images, such as the image exhibited in Figure 3(d).

**3.2 Colorization by NST-CNN model**

To solve the background problem that we meet in EE-CNN model, a new pixel-to-pixel neural style transfer convolutional neural network (NST-CNN) model based on the EE-CNN model was developed. Firstly, some observation results from microscopy images, especially obtained by transmission/scanning electron microscopy, only have the main objects with black/white background. This is due to the usage of mesh grid to support the nanostructure samples. In this case, the NST-CNN algorithm will only colorize the main objects while the background remains in black, compared with EE-CNN algorithm result in Figure S2(c). In most of the colorizing cases, it might have the requirements that a specific color is required for the main objects. Therefore, in this model a reference picture is selected as the guiding color. For example, the input SEM image of virus (Figure 4(a)) is colorized beautifully with its reference image of bread (Figure 4(b)) as demonstrated in Figure 4(c). The image of flower anthers presented in Figure 4(d) is designed for getting the color from another colorful brain neuron image which also has black background (Figure 4 (e)). The final colorized result revealed the same uniform color as the reference image (Figure 4(f)). Furthermore, it is obvious to know that some objects must represent their real-world representation such as red blood cell which need to have a red color. As a result, red color reference image (Figure 4(h)) is selected. Compared with the results made by EE-CNN, a pure black background colorized picture is obtained, together with specified reference color on main objects (Figure 4((c), (f), (i))).

To check the quality of colorization results by our NST-CNN model, the saturation value and hue value from "*HSV*" color space is also analyzed. The colorized sample in Figure 4(f) is selected as the sample. 3D intensity distribution of saturation reveals a rich saturation value in anther structures (Figure 4(j)), but with the only value of 255 (black) in background. Hue value can be typically represented the properties of color. The distribution of hue intensity from our colorized sample (Figure 4(k)) exhibits an uniform value of one color, with only some tiny color noise inside. Furthermore, we made a comparison with previous general proposed model [19]. For the first row of grayscale images (Figure S3 ((a)-(d))), the second row and third row represent the results of previous proposed approach (Figure S3 ((a1)-(d1))) and our methods (Figure S3 ((a2)-(d2))), respectively. Comparatively, it is clear show that some of the background is colorized with color (Figure S3 (a1)). In addition, the individual object is colorized by the unnecessary multi-color (Figure S3 ((b1), (c1), (d1))), since they are the same object have the same properties.

In another case, the majority of samples might have a substrate under the primary object. These substrates can be the surface on which the material deposited on the wafer or the specially fabricated micro/nano structures on substrates. During the imaging stage, the microscopy images will cover both objects and substrates, for example, the chromosome deposited in the surface of substrates (Figure 5((a), (b))). When colorizing these images, the algorithm divides the primary object (orange arrows in Figure 5(a) and 5(b)) as foreground and the substrate (blue arrows) on the image as background. By our NST-CNN method, the colorized results exhibited a distinguishable double color images in Figure 5((c), (d)) with a reference picture (randomly selected), which is not skilled by our EE-CNN model.

Our NST-CNN model can also show improvement if we manually pre-process the image. Figure 6(c) is a microscopy image result from the SEM microscope which have two layers of visible objects (bedbug in the fibre) in the input image. Adaptive-thresholding is manipulated in the input image so that it can separate the two objects in the image. After the thresholding, two reference images are applied (Figure 6((a), (b))) for each separated content to get a colorful result (Figure 6(d)). The Hue value of analysed result exhibits a balanced colorization ability (Figure S4). However, the method shows improvement on limited images. This is because the adaptive thresholding cannot be well enough to identify and group objects in an image.

Finally, a quantitative metric is proposed to measure the performance of general colorization task which based on survey mechanism. First, 16 images are randomly selected from our dataset and 16 images from the predictions of our model. Then, the 32 images mixed randomly. Participants are invited to select 16 images which they consider as the predictions from our model. After submitting the result, the accuracy ($Y_i$ and $Y_i'$) of the selection of participant is calculated. The higher of the accuracy means the better performance of our model. The results are shown in Table S1. It can be seen that the NST-CNN method is slightly better than the EE-CNN method since we give a specific reference image for each input SEM-image. The accuracies of two methods are both closed to 0.5, which means a human cannot distinguish which one is select from human-printed and which one is the prediction from our model.

# 4 Conclusion

To summarise, in this work, we have shown that fully implemented two deep learning architectures can able to colorize nanoscale microscopy images with the natural and designed color. With the help of pre-trained Inception ResNetV2 model, most grayscale images without substrate could be successfully colorized with proper color by our EE-CNN algorithm. Besides, holistically nested edge detection is applied to optimise the approach to high contrast images. In order to colorize the images with a black background or substrate, we further develop the NST-CNN model based on EE-CNN. By giving reference images, our methods perform well colorization ability with the separation of the substrate. Furthermore, multi-objects images could also be well colorized with different color by giving a corresponding number of reference images. By using generative models and extending our dataset, we believed that it can improve the precision of colorization in the future. Overall, we have provided colorization methods with neural networks that can precisely transfer the grayscale images to colorful and attractive images without sacrificing their original appearance, which provides superior help to scientists or other fields such as arts.

**Acknowledgements**

This work was supported by the National Natural Science Foundation of China (Nos. 11804174), Natural Science Foundation of Ningbo (2018A610319) and K. C. Wong Magna

Fund in Ningbo University. We also thank Nvidia for donating NVIDIA DGX-1 used for this research.

**Electronic Supplementary Material:** Supplementary material (grayscale images links from websites; Results from EE-CNN model; Comparison results with the methods from Zhang et al.; The Hue value analysed results from Figure 6) is available in the online version of this article.

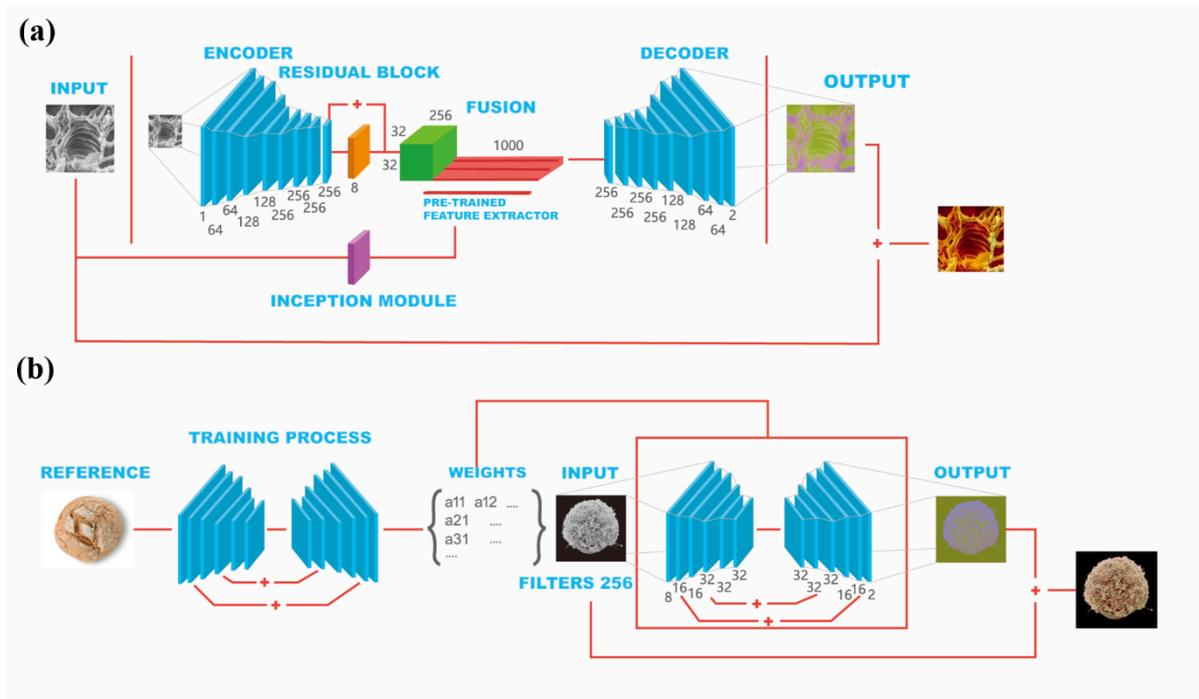

**Figure 1** The overview of proposed mode: (a) End-to-end CNN (EE-CNN) model with encoder-decoder structure. The model has encoder to obtain important features, inception module to extract features from real-world image and finally the decoder uses the features from encoder and inception module to predict the output. (b) Pixel-to-pixel style transfer-based CNN (NST-CNN) model. The model architecture takes two inputs: one image as reference and grayscale microscopy image as a style image (in the middle of the cartoon). The model has same CNN structure as EE-CNN model, but the full network architecture is like neural style transfer. The model trains on one image then it saves the weights at the inference time. Finally, transfer of the trained weight color to the microscopy images.

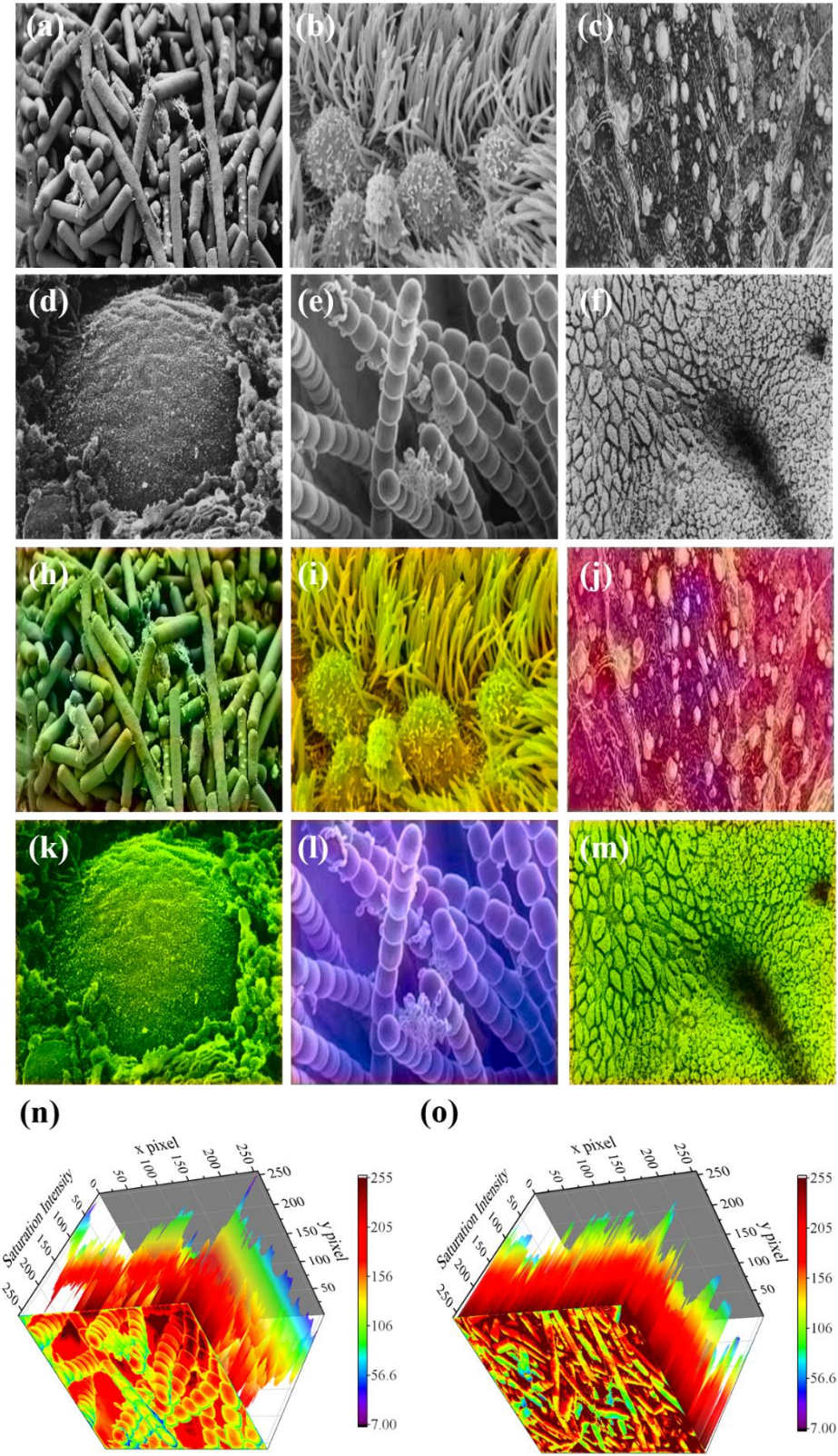

**Figure 2** Results from EE-CNN model and color analysis of EE-CNN model results in "*HSV*" color space. Input grayscale images of bacillus (a), cilia (b), wood surface (c), nuclear

membrane (d), stamens (e), leaf surface (f). Together with their colorized results ((h)-(m)), respectively. (n) 3D intensity distribution of saturation value from figure 2(l) and (o) 3D intensity distribution of saturation value from Figure 2(h). The gray surface in top side of each figure represent the saturation intensity distribution of grayscale image from Figure 2(e) and Figure 2(a).

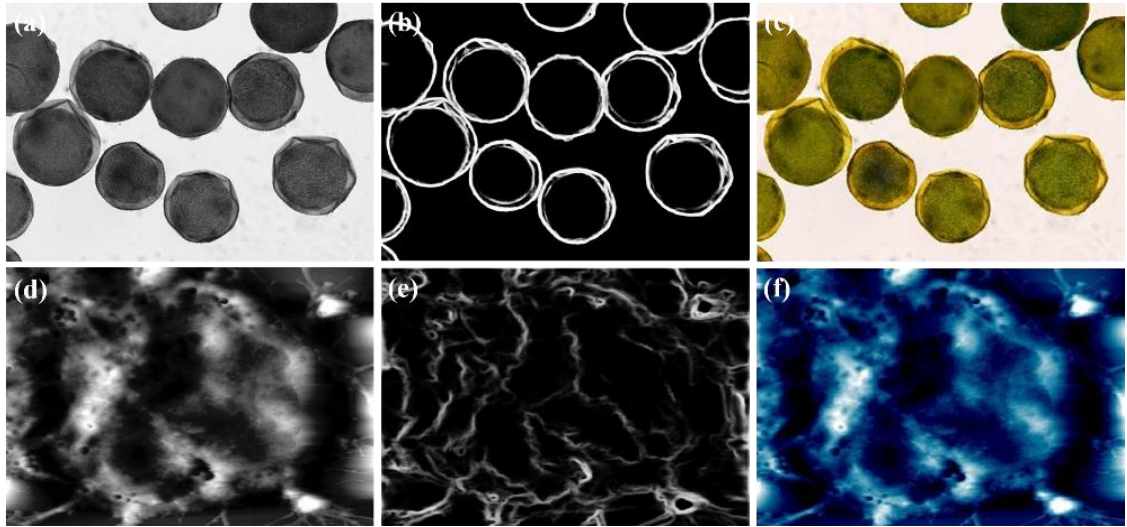

**Figure 3** Holistically nested edge detection based colorization. (a), (d) Input images. (b), (e) The detected edge images from edge detection algorithm and the colorized results ((c), (f)) finally.

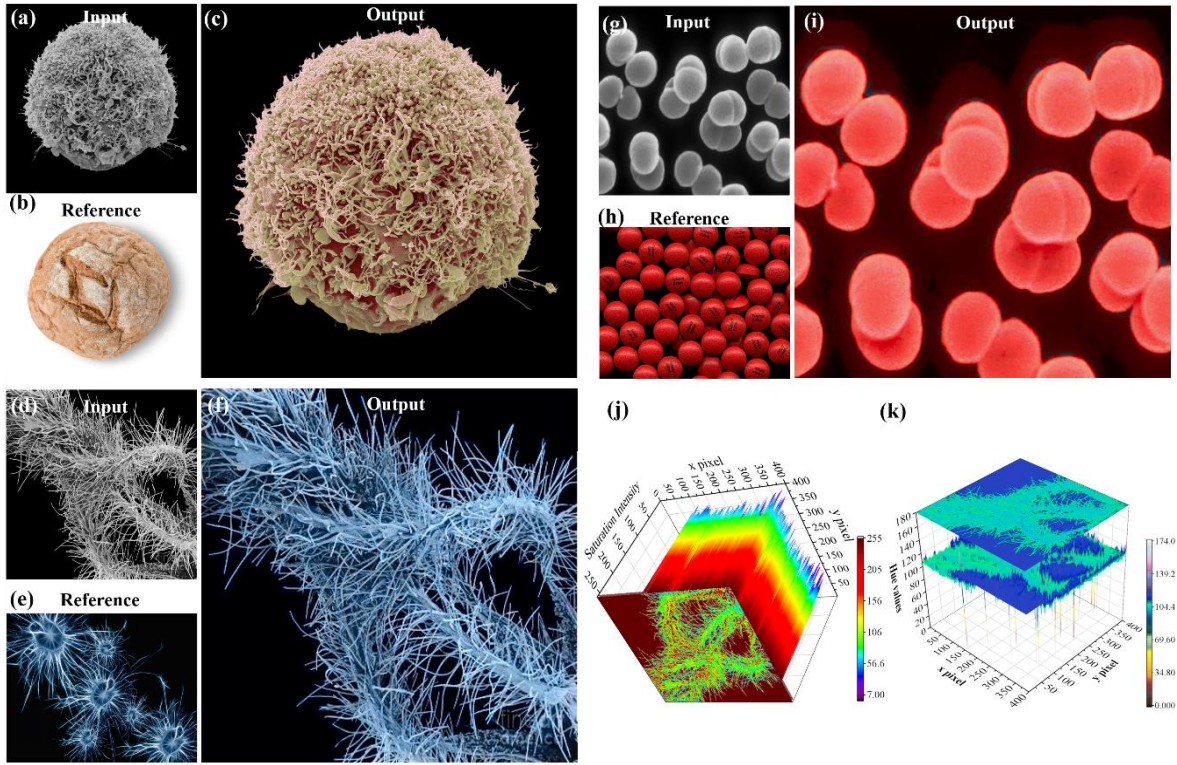

**Figure 4** Colorizing results for the images with a black background and color analysis of our NST-CNN model results in "*HSV*" color space. (a) Input grayscale virus SEM image, reference image of (b) bread and (c) colorized result accordingly. (d) Input grayscale anthers SEM image. (e) Reference image of blue colorized neuron cell, and (f) the result image. (g) Input sample SEM image of red blood cells, (h) red tablets picture as a reference, and (i) colorization result. (j) 3D intensity distribution of saturation value and hue value (k) from Figure 4(f).

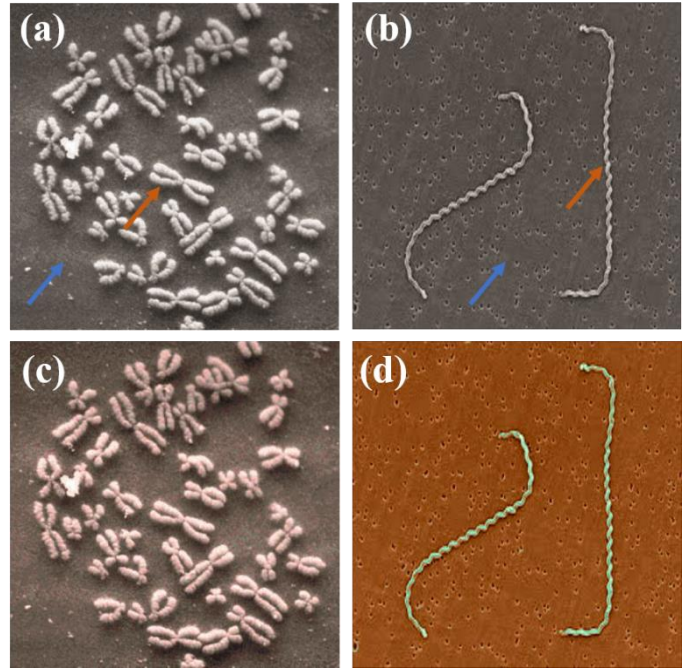

**Figure 5** Colorizing results of grayscale images with main objects and substrates. (a) Grayscale image of chromosome on substrate and (b) Double helix DNA on substrate. Colorization results by NST-CNN algorithm with (c) pink DNA and (d) green DNA fiber.

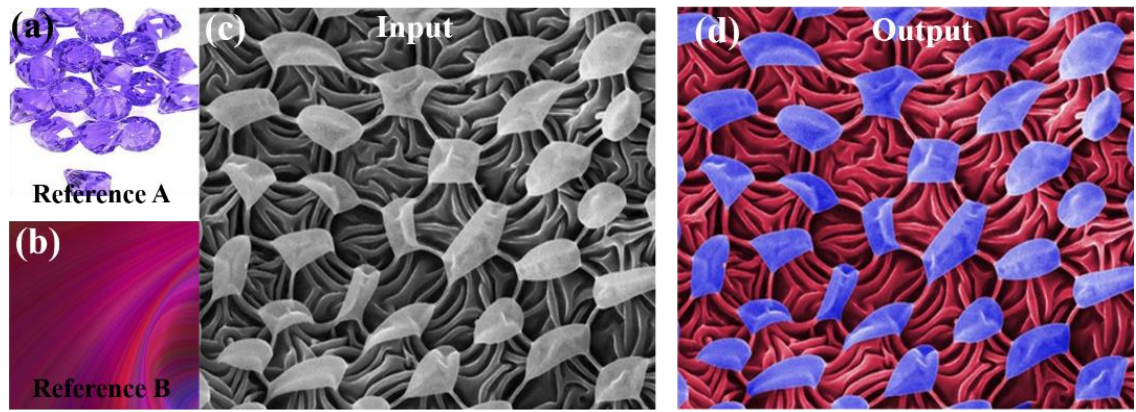

**Figure 6** Colorizing results with manual pre-processing for an image with two contents. (a) and (b) Two reference images. (c) Grayscale image and (d) result image.